%% file: Paper-0039.tex
\documentclass[runningheads]{llncs}
\usepackage[T1]{fontenc}
\input{preamble}
\usepackage{graphicx}

\usepackage{color}

\begin{document}


\title{Auditing Significance, Metric Choice, and Demographic Fairness in Medical AI Challenges}

\titlerunning{Medical AI Challenges}
%

\author{Ariel Lubonja \and
Pedro R. A. S. Bassi \and
Wenxuan Li \and
Hualin Qiao \and \\
Randal Burns \and
Alan L. Yuille \and
Zongwei Zhou~\Envelope
}

\authorrunning{A. Lubonja et al.}
\institute{
Johns Hopkins University \\
\href{mailto:zzhou82@jh.edu} {\textsc{zzhou82@jh.edu}}
}

    
\maketitle              

\begin{abstract}

Open challenges have become the \textit{de facto} standard for comparative ranking of medical AI methods. Despite their importance, medical AI leaderboards exhibit three persistent limitations: (1) score gaps are rarely tested for statistical significance, so rank stability is unknown; (2) single averaged metrics are applied to every organ, hiding clinically important boundary errors; (3) performance across intersecting demographics is seldom reported, masking fairness and equity gaps. We introduce \method, an open‑source toolkit that seeks to address these limitations. \method\ (1) computes pair‑wise significance maps that show the nnU‑Net family outperforms Vision‑Language and MONAI submissions with high statistical certainty; (2) recomputes leaderboards with organ‑appropriate metrics, reversing the order of the top four models when Dice is replaced by NSD for tubular structures; and (3) audits intersectional fairness, revealing that more than half of the MONAI‑based entries have the largest gender‑race discrepancy on our proprietary Johns Hopkins Hospital dataset. The \href{https://github.com/ariellubonja/RankInsight-Medical-Segmentation-Statistics}{\method}\footnote{\url{https://github.com/ariellubonja/RankInsight-Medical-Segmentation-Statistics}} toolkit is publicly released and can be directly applied to past, ongoing, and future challenges. It enables organizers and participants to publish rankings that are statistically sound, clinically meaningful, and demographically fair.

\end{abstract}


\section{Introduction}\label{sec:introduction}

Open challenges have become the cornerstone for evaluating medical AI methods \cite{FLARE23-ma2024automaticorganpancancersegmentation,bassi2024touchstone,wiesenfarth2021methods}. Organizers release a carefully curated dataset, define a clinical task (e.g., multi‑organ segmentation), specify evaluation metrics, and maintain a public leaderboard \cite{bilic2019liver,antonelli2022medical,heller2019kits19,chen2025vision,bassi2025radgpt,zhang2025automated}. 


However, most leaderboards exhibit three critical limitations: \textbf{(1) No statistical analysis.} Many challenges' participants---top methods especially---differ by only $\leq 0.1$ of a metric  \cite{bassi2024touchstone,FLARE23-ma2024automaticorganpancancersegmentation,luo2024rethinking}. A small change in score leads to a large change in leaderboard position, obscuring whether rankings are statistically significant, generalizable or reproducible \cite{wiesenfarth2021methods,reinke2023challenge}. \textbf{(2) Limited metric suitability.} Reliance on a single metric such as the Dice Similarity Coefficient (DSC), though intuitive for leaderboards, may not reflect clinical reality \cite{reinke2024understanding}. For example, DSC rewards high area overlap, but in various clinical applications, precise boundary segmentation is more important.
\textbf{(3) Neglected demographic fairness.} If not carefully controlled, challenges' datasets may unintentionally promote methods that focus on majority patients (e.g. white, older) \cite{larrazabal2020gender,li2024text}, failing to maintain performance parity on diverse and intersectional populations, e.g., young African American Female patients \cite{seyyed2021underdiagnosis,lekadir2025future,seyyed2020chexclusion,li2025scalemai}.

To address these limitations, the community has developed several toolkits \cite{wiesenfarth2021methods,reinke2024understanding}, however, though published in respected venues and well-cited, adoption by challenge organizers remains low. We believe the reason for this low adoption is that the toolkits are difficult to use\textemdash understanding methods such as Wilcoxon test with Bonferroni correction requires specialized statistical knowledge\textemdash and results interpretation is not straightforward. We aim to address this by contributing an \textbf{open-source toolkit}, called \method, that automates the creation of: (1) statistical significance maps through pairwise hypothesis tests, (2) choice of the appropriate metric based on the target class, and (3) demographic parity measures and per-group performance highlights.  In hope of pushing adoption, we design our toolkit to work with existing challenge data, without requiring manual \& time-consuming preprocessing. To illustrate it's usefulness, we applied this toolkit to the recent Touchstone challenge \cite{bassi2024touchstone}  and observed:

\begin{enumerate}

    \item \textit{Statistical Significance of Rankings (\S\ref{method:significance-maps})}: We apply pairwise statistical significance tests between methods to determine the confidence with which one method can be said to perform better than another (Fig. \ref{fig:significance-heatmap}). We reveal the nnU-Net framework confidently outperforms the Vision-Language and MONAI frameworks, while within-family tests are mostly inconclusive.
    
    \item \textit{Rankings Based on Appropriate Metrics (\S\ref{method:metrics})}: We analyze the change in leaderboard rankings caused by the usage of DSC metrics for blob-like, area-dependent organs and NSD for boundary-critical, elongated or tubular organs (Fig. \ref{fig:per-organ-heatmap}). We aggregate these changes to provide a single, unified score.
    
    \item \textit{Demographic Group \& Intersectional Performance (\S\ref{method:demographics})}: We evaluate the methods' Demographic Parity Difference \cite{chen2023algorithmic}, and reveal the most equitable (MedFormer \cite{gao2022data} and U-Net CLIP \cite{liu2023clip,liu2024universal,tang2024efficient,zhang2023continual}) and least-equitable methods (MONAI-based UNETR \cite{hatamizadeh2022unetr} and UNEST \cite{yu2023unest}) in Fig. \ref{fig:intersectional-dpd}.
    
\end{enumerate}

\section{Related Work}

Several recent works examine the statistical significance and validation shortcomings in challenge leaderboards. Antonelli \& O'Reiller et al. show that very few challenges provide deeper statistical analysis or reproducibility checks \cite{antonelli2022medical,oreiller2022head}, and they provide no demographic auditing. Reinke~\etal~\cite{isensee2024nnu} further show that claims of state of the art methods often lose their lead when accounting for uneven baselines or validation shortcomings. The problem extends beyond challenges: in a recent review of 223 segmentation papers in MICCAI 2023, Christodoulou~\etal~\cite{christodoulou2024confidence} found that 50\% reported no measure of variability, and only one paper (<0.5\%) provided confidence intervals.

Efforts to detect and address these shortcomings have been introduced. Wiesenfarth~\etal ~\cite{wiesenfarth2021methods} advocated for and implemented a framework that contains functions such as ranking stability analysis, significance maps, statistical tests, and plotting techniques such as box-plots and violin plots. Detecting  and presenting methods' failure modes has been addressed in Roß~\etal~\cite{ross2023beyond}, testing input corruptions such as object occlusion, underexposure and presence of unexpected objects. The question of what distinguishes top-ranked methods, and how they utilize the above insights is revealed in Eisenmann~\etal~\cite{eisenmann2023winner}. 

Reinke~\etal~\cite{reinke2024understanding} outline characteristics and pitfalls of segmentation metrics, highlighting the non-triviality of metric choice and dangers of relying on popular choices. Müller~\etal~\cite{muller2022towards} highlight the implications of region of interest size on various common metrics.

Studies in AI fairness and under-diagnosis in population subgroups have been performed by gender \cite{larrazabal2020gender,seyyed2021underdiagnosis,li2025pants,chen2024analyzing}, race or socioeconomic background \cite{seyyed2021underdiagnosis,seyyed2020chexclusion}, revealing concerning diagnostic inequity by state of the art MAI algorithms.  Gichoya~\etal~\cite{gichoya2022ai} showed that MAI models can infer a person's race from CT imaging\textemdash even when radiologists cannot, and no correlation between race and diagnosis is known. Metrics such as Demographic Parity Difference,  and Predictive Parity \cite{chen2023algorithmic}, Equalized Odds \cite{hardt2016equality}, Predictive Parity \cite{chouldechova2017fair}, and True Positive Rate disparity \cite{seyyed2020chexclusion} have been proposed to address these discrepancies.

\section{Method}\label{sec:method}

\subsection{Ranking Significance Maps}\label{method:significance-maps}

We first address the issue of rank uncertainty by performing exhaustive pairwise hypothesis tests between competing methods. For each pair, we conduct a paired one-sided Wilcoxon signed-rank test on their per-sample scores\footnote{We choose the Wilcoxon test as a non-parametric alternative to the paired $t$-test because metric distributions often violate normality assumptions; Wilcoxon operates on the ranks of score differences and is robust to outliers.}. Given $M$ methods, we have $M(M-1)/2$ pairwise comparisons. To control the family-wise error rate, we apply a Bonferroni correction (or equivalently, test at a stricter $\alpha$). We set a default overall $\alpha=0.05$. Rather than force a binary significant/not-significant decision, we output and visualize confidence levels. Specifically, we plot a $M\times M$ significance map where each cell is color-coded to indicate the confidence that the row method outperforms the column method. For example, a cell might be colored green if the row method $m_r$ is significantly better than the column method $m_c$, or gray if $m_r$ is not better than $m_c$: either the difference is statistically indistinguishable, or $m_c$ outperforms $m_r$. This yields an intuitive heatmap where a row with many solid green cells indicates a method that outperforms many others with high certainty. Conversely, a method $m$ that has mostly gray cells in its row is highly unlikely to outperform others.



\subsection{Ranking Based on Appropriate Metrics} \label{method:metrics}

\begin{figure}
    \centering
    \includegraphics[width=.5\linewidth]{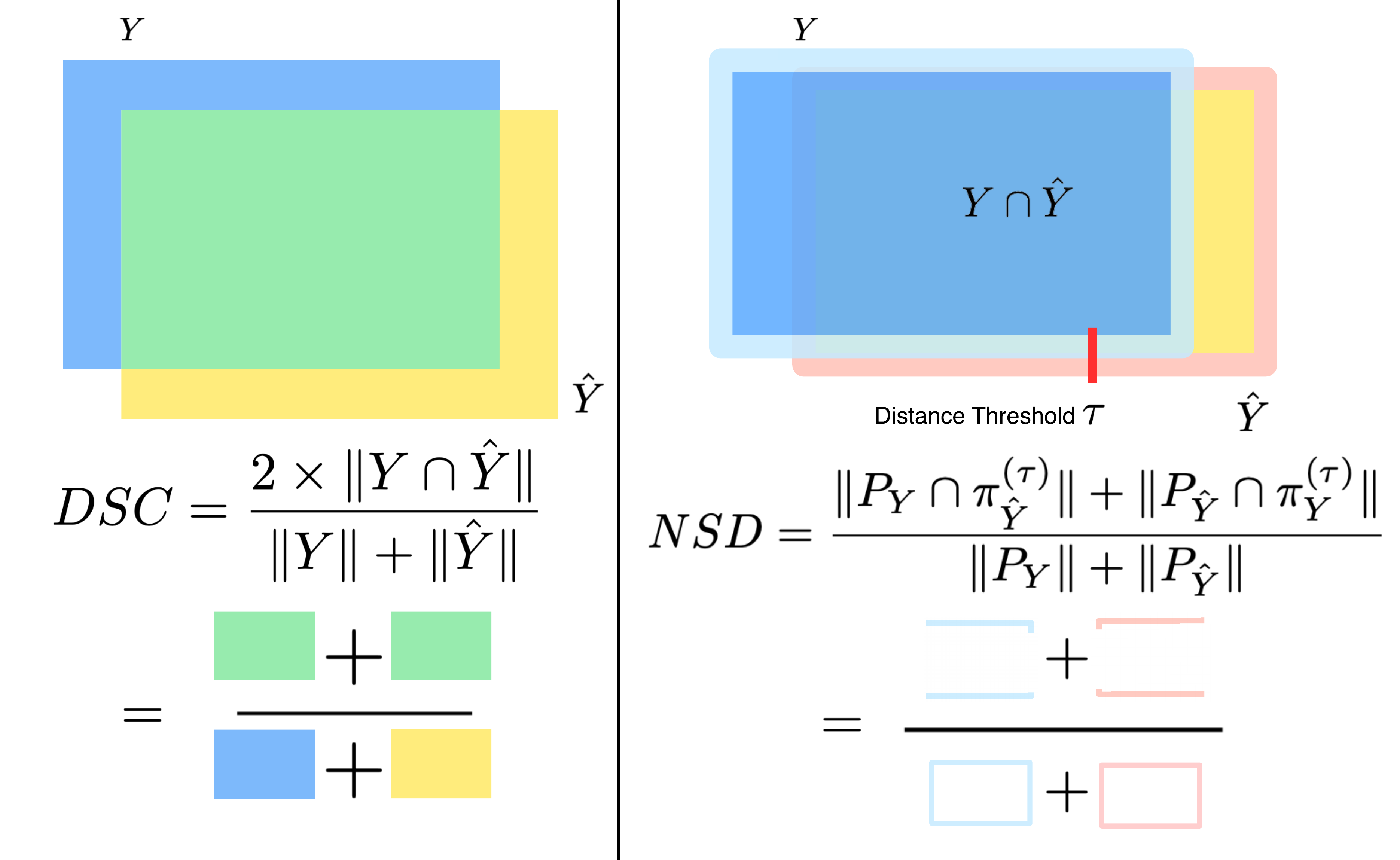}
    \caption{Illustration of calculation of DSC and NSD. DSC (left) is an area overlap-based metric, favoring area agreement. NSD (right) is a perimeter-based metric, calculating the proportion of an object's boundary ($P_Y, P_{\hat{Y}}$) that lies within the other object's frontier region $\pi_Y^\tau$. $\tau$ is the distance threshold that provides leniency for non-perfect boundary matches.}
    \label{fig:dsc-nsd-illustration}
\end{figure}

Different anatomical structures pose different segmentation challenges, so no single metric suits all organs. To make leaderboards clinically meaningful, we adopt an organ-specific metric mapping. Large, solid organs (e.g., liver, spleen, tumors) are evaluated using Dice Similarity Coefficient (DSC), which emphasizes volume overlap. In contrast, thin or tubular structures (e.g., vessels, ducts, spinal cord) are assessed using Normalized Surface Distance (NSD)\footnote{NSD measures the proportion of a predicted boundary that lies within a set tolerance of the ground truth boundary, penalizing misalignments and gaps. The tolerance (e.g., 1–2 mm) is organ-specific and user-adjustable.}. Fig.~\ref{fig:dsc-nsd-illustration} illustrates why DSC may miss boundary errors and how NSD better captures them.

\method\ supports either a built-in dictionary of organ-to-metric mappings or custom user-defined ones. For example, heart, liver, and kidneys default to DSC, while vessels and airways use NSD \cite{lin2025pixel,liu2025shapekit}. In the TotalSegmentator challenge, we applied this mapping automatically. For each submission, both DSC and NSD are computed, and rankings are generated per organ using the appropriate metric. This yields two parallel leaderboards: one using the original metric and one using the revised, organ-aware mapping. A final report summarizes each method’s per-organ rankings under both metrics.

\subsection{Demographic Discrepancy Analysis}\label{method:demographics}

Finally, our \method\ toolkit includes a fairness auditing module to evaluate model performance across demographic and protected subgroups. Given metadata such as race, sex, age group, or scanner type, we assess whether model accuracy is consistent across these groups or if performance gaps indicate potential bias. We use the Demographic Parity Difference (DPD) as a fairness metric, where demographic parity implies that all groups should have equal rates of positive outcomes.

For segmentation, we define a \textit{successful outcome} as achieving a quality threshold $t$ (e.g., Dice > 0.8 for the main organ). DPD is then the difference in success rates between two groups. For example, if a model meets the threshold in 90\% of male cases and 80\% of female cases, $DPD = 0.10$ indicates a disparity. We compute DPD for each attribute (e.g., sex, race) and also for intersections (e.g., “Black female” vs. “White male”).

\smallskip\noindent\textbf{Intersectional Analysis:} We automatically identify the worst-case performance gap across all subgroup combinations (e.g., \{male, female\} $\times$ \{White, Black, …\} $\times$ \{scanner A, B, …\}). Due to data sparsity, not all intersections are reported; we require at least $n \geq 40$ samples per group. Users can set a threshold $\tau$ (e.g., 0.10), and the toolkit flags subgroup pairs with $DPD > \tau$ as significant disparities. Results are visualized via bar plots, tables, or heatmaps for easy interpretation.

\section{Results}\label{sec:results}
    
\subsection{Statistical Significance Ranking on TotalSegmentator}\label{results:significance-maps}

\begin{figure}[t]
    \centering
    \includegraphics[width=0.95\linewidth]{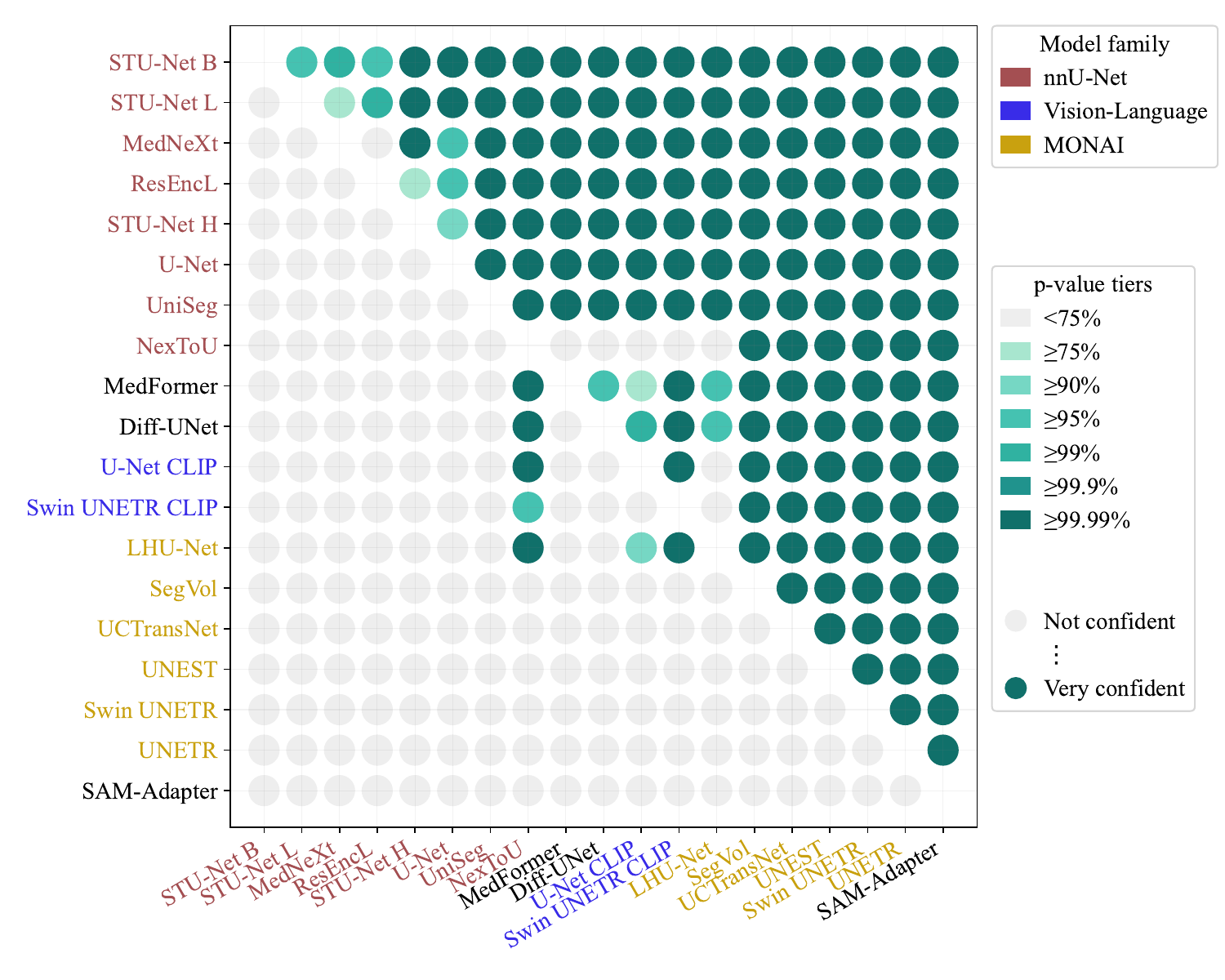}
    \caption{Pairwise significance map of model rankings on \textit{TotalSegmentator}.
    Dark–green cells denote comparisons where the row method is ranked higher than the column method at $p<0.001$. Progressively lighter shades indicate lower, but still meaningful, confidence tiers.  
    nnU-Net-based submissions (dark red) form a clear top tier that consistently outperforms the remaining models (upper-right quadrant), whereas MONAI-based models (gold) appear to underperform (lower right quadrant).
    Intra-model family comparisons (close to the diagonal) contain many gray cells and light-green cells, cautioning their comparative ranking is less certain.
    }
    \label{fig:significance-heatmap}
\end{figure}

\noindent\textbf{Experiment Protocol.} The 19 methods were run against TotalSegmentator, and their per-organ DSC score was calculated and averaged. Missing predictions were discarded. Given these predictions, we performed a pair-wise, one-sided Wilcoxon test for all 19 submitted methods. Fig.~\ref{fig:significance-heatmap} shows the models, grouped into color-coded families where applicable, to highlight framework similarities. We threshold the $p$-value at multiple levels to provide intuitive confidence tiers (gray to dark green). Axes are sorted by aggregate model family performance, then individual model performance within each family.

\smallskip\noindent\textbf{nn-Unet Model Family demonstrates clearly superior performance.} Fig.\ref{fig:significance-heatmap} shows that nearly all nnU-Net variants (dark red) significantly outperform other methods ($p<0.001$), forming a strong dark-green block in the upper-right quadrant. The only exception is NexToU~\cite{shi2023nextou}. In contrast, the lower-left quadrant is mostly gray, indicating that no other model family beats nnU-Net with statistical significance. These results highlight the robustness of nnU-Net’s design—large patch sizes, heavy test-time augmentation, and tailored post-processing. A second tier includes the Vision–Language models (blue)—VL-Swin UNETR-CLIP~\cite{liu2023clip}, VL-U-Net-CLIP~\cite{liu2023clip}, Diff-UNet~\cite{XING2025diffunet}, and MedFormer~\cite{gao2022data}. These models consistently outperform the MONAI-based methods (gold), but are in turn clearly outperformed by nnU-Net models. MONAI-based models perform weakest overall, failing most pairwise comparisons except against SAM-Adapter~\cite{gu2024build}. Together, the results suggest a ranking of model families: (1) nnU-Net, (2) Vision–Language and Diff-UNet/MedFormer, and (3) MONAI.

\smallskip\noindent\textbf{Within model family ranking is ambiguous.}
Within the nnU-Net group, ten comparisons are low-confidence, including among the top-3. As a result, we cannot draw a definitive top-3 conclusion from this dataset. Surprisingly, the three sizes of STU-Net \cite{huang2023stu} models are performing in reverse order of size: the smallest (Base) model appears first, followed by Large and further behind the Huge scale version. A similar, stronger pattern is observed within the Vision–Language and independent models, in the center of the matrix. Although the weakest model among them is clear, Swin UNETR Clip, we cannot draw reliable conclusions on which among MedFormer, Diff-UNet or U-Net CLIP is better. The MONAI framework models show a clear pattern, with LHU-Net being the comfortable winner, and a solid ranking being formed.

\subsection{Appropriate Metrics Applied to TotalSegmentator Rankings} 

\begin{figure}[t]
    \centering
    \includegraphics[width=1\linewidth]{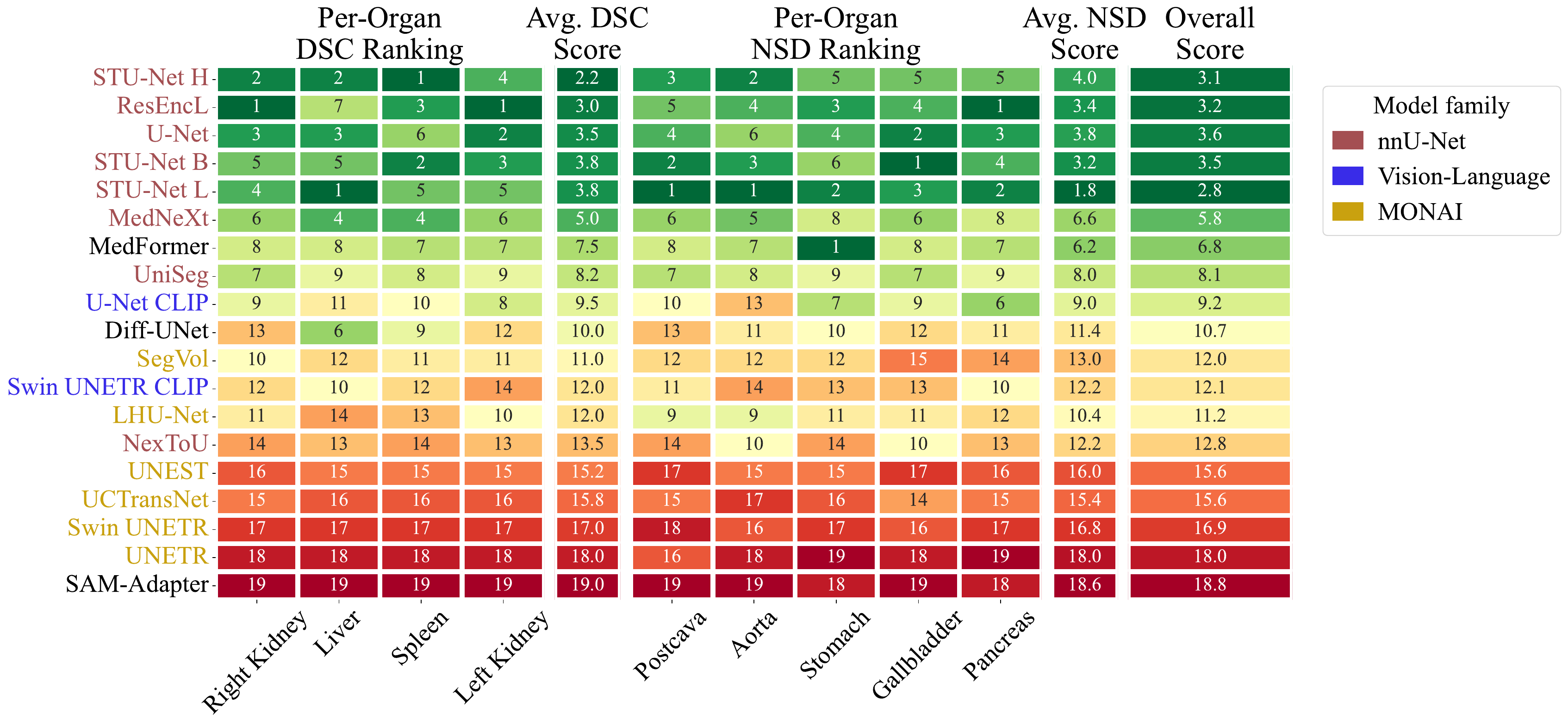}
    \caption{Relative ranking of methods based on their per-organ performance, using DSC for blob-like organs and NSD for tubular, elongated organs. Average Scores for each metric, and Overall (unweighted) score are calculated. Table is sorted by Average DSC score. Average NSD Score shows a reversal of the model rankings for the top-performing, nnU-Net-based methods. }
    \label{fig:per-organ-heatmap}
\end{figure}

A common challenge approach is to present the models ordered by their average DSC or NSD. It is interesting to see how the Touchstone model rankings would change if DSC and NSD were applied selectively to their respective organs. Fig.~\ref{fig:per-organ-heatmap} presents the per-organ ranking, as well as a DSC, NSD and Overall averaging score. We sorted our table by DSC since it is the most common metric used to rank models in challenges. When placed side-by-side with the NSD results, the ranking of the top-5 models flips.

Which model should you pick? Assuming both DSC and NSD performance are equally important to you, we introduce an Overall Score, which can be retroactively applied to challenge results. In our final Overall Score, the top models remain top, but the within top-5 ranking changes significantly, mirroring our \S\ref{results:significance-maps} results. An unlikely performer is MedFormer, which goes from mid-tier to state of the art for Stomach segmentation. The cause of the rank changes is the difference in DSC and NSD calculations, as illustrated in Fig. \ref{fig:dsc-nsd-illustration}.

\subsection{Intersectional Performance and Demographic Fairness}

\begin{figure}[t]
    \centering
    \includegraphics[width=1\linewidth]{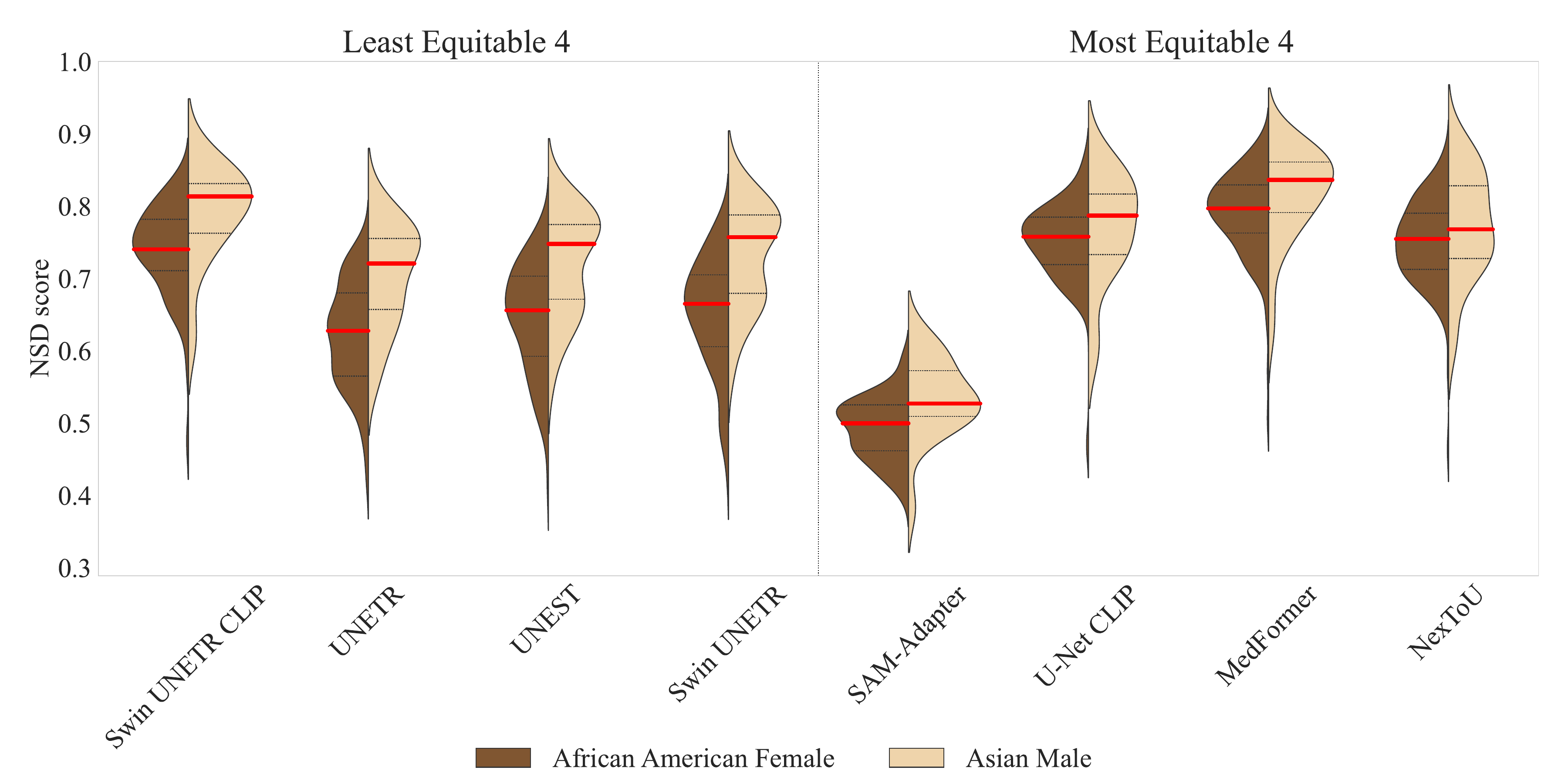}
    \caption{Split Violin plots for least (left) and most (right) equitable models tested, on an Intersectional view of Gender-Race. On each split violin plot, left side is a Kernel Density Estimate of the model's NSD Score on African American Female patients, and right on Asian Male patients. Mean NSD score is denoted by a solid red line. Demographic Parity Difference calculates the difference of means. Three UNETR-based architectures (UNETR, Swin UNETR and Swin UNETR CLIP) and UNEST have the biggest gender-race discrepancy. Two nnU-net-based models (NexToU, MedNeXt) one VL-Based (U-Net CLIP) and SAM-Adapter were most equitable. Most equitable does not imply best, as highlighted by SAM-Adapter.}
    \label{fig:intersectional-dpd}
\end{figure}

Challenges rarely report model performance by any single demographic feature, much less intersections of two or more protected features. TotalSegmentator does not include this information, but our proprietary Johns Hopkins Hospital (JHH) dataset does. Our toolkit iterates over the combinations of protected features, finding the biggest discrepancies, in our case between Asian Males and African American Females. Ranking the models by their DPD score, we present the score distribution for the 4 least (left) and most (right) equitable models (Fig.~\ref{fig:intersectional-dpd}).

\smallskip\noindent\textbf{None of the top-5 models from our above results appear among in most equitable list.} The worst-performing nnU-net-based model, NexToU, tops our equity list. The highest-ranked model in our top-4 is MedFormer, is ranked 7 in our Overall Score in our test, highlighting a potential trade-off in model choice for underrepresented groups. SAM-Adapter ranks very highly in DPD, despite underperforming every other model\textemdash the most equitable models are not necessarily strong performers. Three UNETR-based, and UNEST show highest inequity, with a mean NSD difference of 0.07--0.12.

\section{Conclusion}\label{sec:conclusion}

We presented a streamlined, open-source toolkit that addresses the common pitfalls of leaderboard-based evaluations in medical AI challenges. By integrating (1) pairwise statistical significance tests, (2) context-appropriate metrics (like NSD for boundary-sensitive structures), and (3) intersectional demographic analyses, we enable a richer, more transparent assessment of competing methods. Our empirical results demonstrate that small numerical differences between methods rarely carry statistically significant weight, unless a lot of data is used, and that single-metric rankings (e.g., purely DSC-based) can overlook crucial shape or boundary errors. Further, we find that average performance can mask large subgroup disparities. We hope this toolkit will encourage challenge organizers to go beyond table-based reporting to include distributions and insights into the methods' performance. Inclusion of such reports would promote the development of robust, clinically meaningful, and fair medical AI methods.

\newpage
\begin{credits}
\subsubsection{\ackname} This work was supported by the Lustgarten Foundation for Pancreatic Cancer Research and the National Institutes of Health (NIH) under Award Number R01EB037669. We would like to thank the Johns Hopkins Research IT team in \href{https://researchit.jhu.edu/}{IT@JH} for their support and infrastructure resources where some of these analyses were conducted; especially \href{https://researchit.jhu.edu/research-hpc/}{DISCOVERY HPC}.

\subsubsection{\discintname}
The authors declare no competing interests.
\end{credits}

%
%
%
\bibliographystyle{splncs04}
\bibliography{Paper-0039,zzhou}

\end{document}

%% file: preamble.tex
\usepackage{float}
\usepackage{pifont}
\usepackage{footnote}
\usepackage{enumitem}
\usepackage{bm}
\usepackage{arydshln}
\usepackage{booktabs}
\usepackage{multicol}
\usepackage{multirow}
\usepackage{color}
\usepackage{xcolor}     
\usepackage{colortbl}
\usepackage{soul}
\usepackage{bbding}
\usepackage{makecell}
\usepackage{mathtools}
\usepackage{imakeidx}
\usepackage{amssymb}
\usepackage{graphicx}
\usepackage{amsmath}
\usepackage{threeparttable}
\definecolor{citecolor}{HTML}{0000FF}
\definecolor{linkcolor}{HTML}{0000FF}
\usepackage[colorlinks,
            anchorcolor=red,
            citecolor=citecolor, 
            linkcolor=linkcolor,
            urlcolor=citecolor
            ]{hyperref}          
\makeindex
\usepackage{arydshln}
\usepackage{lipsum}
\usepackage[toc]{multitoc}
\usepackage[edges]{forest}
\usepackage[normalem]{ulem}

\usepackage{bbding}
\usepackage[most]{tcolorbox}

\usepackage{algorithm}
\usepackage{algorithmic}

\usepackage{minitoc}
\usepackage[toc,page,header]{appendix}

\definecolor{orchid}{rgb}{0.85, 0.44, 0.84}
\definecolor{rubinred}{rgb}{0.82, 0.0, 0.28}
\definecolor{flagship}{rgb}{0.93, 0.06, 0.41}
\definecolor{radiologist}{rgb}{0.50, 0.50, 1}

\newcommand{\etal}{\mbox{et al.}}

\definecolor{YT}{HTML}{002FA7}

\newcommand{\method}{{\fontfamily{ppl}\selectfont
RankInsight}}

\newcolumntype{P}[1]{>{\centering\arraybackslash}p{#1}}
\newlength\savewidth
  